\documentclass[sigconf,authorversion,nonacm]{acmart}

\AtBeginDocument{%
  \providecommand\BibTeX{{%
    \normalfont B\kern-0.5em{\scshape i\kern-0.25em b}\kern-0.8em\TeX}}}

\acmConference[AAAI '24]{Make sure to enter the correct
  conference title from your rights confirmation emai}{August 05--07,
  2024}{Newport Beach, CA}
\usepackage{mathtools}
\usepackage[ruled, vlined, noend]{algorithm2e}

\begin{document}

\title{PQS (Prune, Quantize, and Sort): Low-Bitwidth Accumulation of Dot Products in Neural Network Computations}

\author{Vikas Natesh}
\email{vnatesh@g.harvard.edu}
\orcid{}
\affiliation{%
  \institution{Harvard University}
  \streetaddress{P.O. Box 1212}
  \city{Cambridge}
  \state{MA}
  \country{USA}
}

\author{H.T. Kung}
\email{kung@harvard.edu}
\orcid{}
\affiliation{%
  \institution{Harvard University}
  \streetaddress{P.O. Box 1212}
  \city{Cambridge}
  \state{MA}
  \country{USA}
}


\begin{abstract}
We present PQS, which uses three techniques together - Prune, Quantize, and Sort - to
achieve low-bitwidth accumulation of dot products in neural network computations.
In conventional quantized (e.g., 8-bit) dot products, partial results are accumulated
into wide (e.g., 32-bit) accumulators to avoid overflows when accumulating
intermediate partial sums. However, such wide accumulators increase memory bandwidth
usage and reduce energy efficiency.
We show that iterative N:M pruning in floating point followed by quantization to 8 (or fewer) bits, and accumulation of partial products in a sorted order ("small to large") allows for accurate, compressed models with short dot product lengths that do not require wide accumulators.
We design, analyze, and implement the PQS algorithm to eliminate accumulation overflows at inference time for several neural networks. Our method offers a 2.5x reduction in accumulator bitwidth while achieving model accuracy on par with floating-point baselines for multiple image classification tasks.

\end{abstract}

\maketitle

\section{Introduction}
To utilize AI for the edge, low-power Internet of Things (IoT) devices and tinyML applications have been employed to perform a variety of tasks, including eye tracking, gesture detection, motion detection, speech recognition, and head movement detection \cite{jorge,siracusa,mcunet,mema}. 
There is a growing demand for efficient implementations of AI with heavily limited memory, bandwidth, and computation power.

To this end, model compression is essential for efficient inference on low-power devices.
Pruning and quantization are two common approaches to compressing neural networks.
Low-power devices for tinyML typically have small local memories \cite{m4} and often lack support for efficient floating-point computation \cite{gap8,siracusa}.
Hence quantization is, by default, a necessity on such systems and most tinyML models are quantized to 8 bits or less.

When performing matrix multiplications with quantized weights and activations, dot products are typically accumulated into 32-bit registers.
Reducing the accumulator bitwidth can reduce bandwidth and energy usage while increasing inference throughput \cite{a2q, wrapnet,henk}.


However, if the sum of partial products overflows the accumulator, its value may be clipped to a finite range.
This introduces numerical errors into the final matrix result that degrade model accuracy and ultimately limit how much we can reduce the accumulator bitwidth.

Prior works have attempted to reduce the incidence of overflows for narrow accumulators through regularization on the loss function  \cite{wrapnet} or by controlling
weight magnitude during training \cite{a2q,henk}. 
While such approaches succeed in reducing overflows, they impose restrictive constraints on weights that may reduce model accuracy \cite{a2q, a2q+, wrapnet}.
Weight magnitude constraints also promote unstructured sparsity in the network \cite{a2q}, which is beneficial for reducing model size. 
However, the efficient implementation of unstructured sparse matrix operations is challenging as non-zero values may be arbitrarily distributed and must be addressed individually using indexing arrays, incurring computation overhead and additional memory footprint.

\begin{figure}[h!]
\centering
\includegraphics[width=1\linewidth]{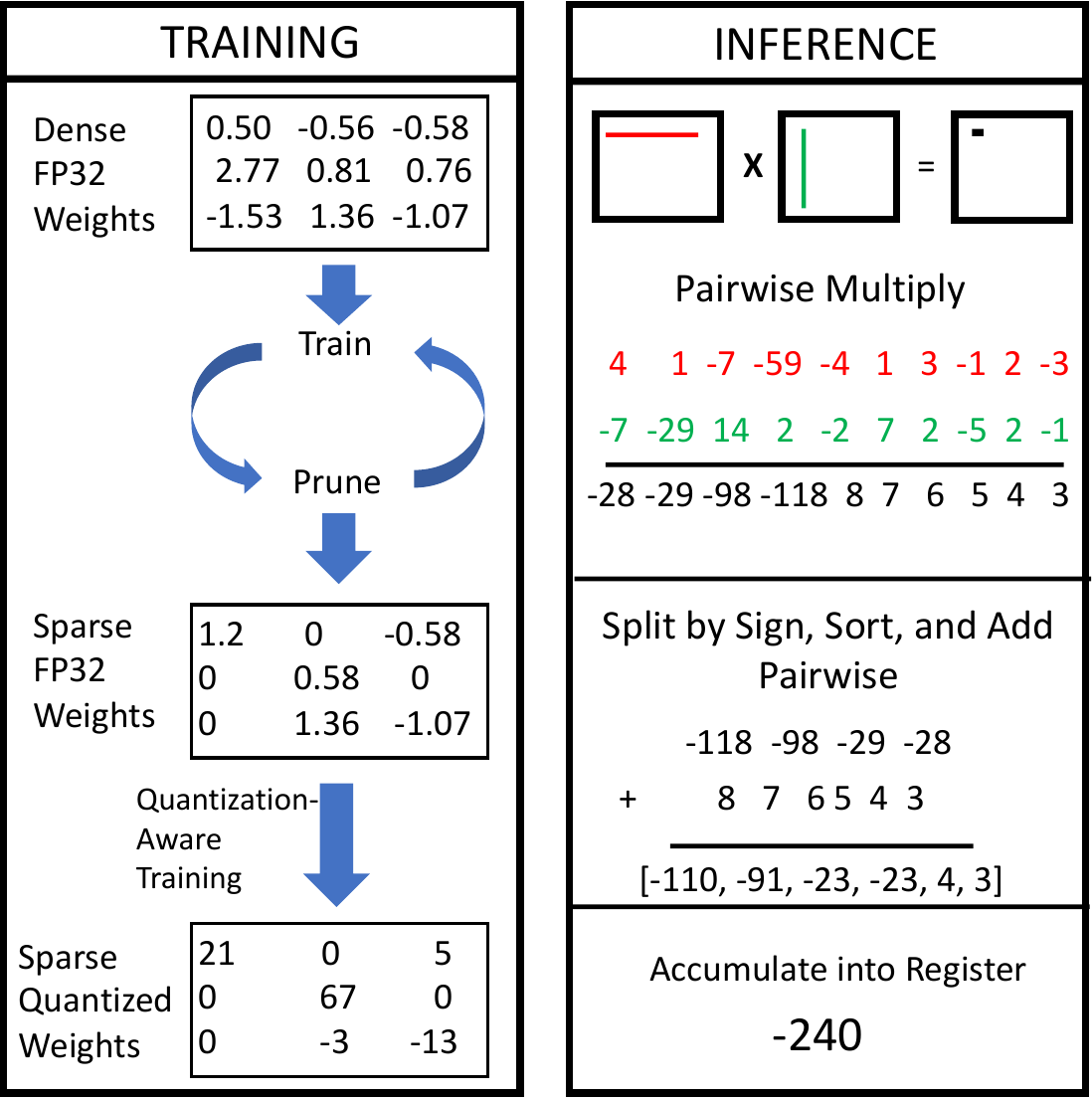}
\caption{
Overview of the proposed PQS (Prune, Quantize, and Sort) framework for enabling low-bitwidth accumulation in quantized neural networks. 
After sparsifying a floating-point model via N:M pruning, we quantize the model using quantization-aware training on the remaining weights. 
During inference, we reduce transient accumulation overflows through our sorted dot product algorithm}
\label{fig:overview}
\end{figure}

We combine pruning, quantization, and a novel sorted dot product algorithm to enable low-precision accumulation in quantized neural networks (QNNs) with minimal accuracy degradation.
We characterize overflows as \textbf{persistent} or \textbf{transient}, depending on whether the final accumulation result overflows or only an intermediate partial accumulation result overflows, respectively.
Instead of directly reducing weight magnitude during training to reduce accumulator magnitude during inference, we show that structured N:M weight pruning can restrict dot product lengths (i.e., the number of partial products) sufficiently to avoid most persistent overflows when using narrow accumulators.
Structured sparsity also yields predictable nonzero distributions in weight matrices and is amenable to software and/or hardware acceleration by skipping unnecessary computation.
We then avoid transient overflows that temporarily arise during inference via a dot product algorithm that sorts partial results before accumulation (Section \ref{sec:sort}).

Combining multiple model compression techniques is non-trivial as the specific order in which techniques are applied can impact accuracy.
We consider two popular training schedules for combining pruning and quantization, namely pruning in floating point (FP32) before quantizing to 8 or fewer bits (P->Q) and quantizing to fewer bits before pruning the quantized weights (Q->P).
In Section \ref{sec:pruning}, we show that floating point weights are a superior pruning signal to quantized weights and P->Q produces models with higher accuracy than Q->P.
Figure \ref{fig:overview} provides a schematic of the training and inference pipeline in PQS.
The novel contributions of this paper are:
\begin{itemize}
    \item Analysis of persistent and transient dot product overflow in quantized neural networks (Section \ref{sec:ovfl})
    \item Sorted dot product algorithm for eliminating transient overflows (Section \ref{sec:sort}).
    \item Combining N:M pruning and quantization to mitigate persistent overflows (Section \ref{sec:pruning}).
    \item Evaluation of the resulting PQS methods in terms of model accuracy and accumulator compression for several neural networks on classification tasks (Section \ref{sec:eval}).
\end{itemize}


\section{Background}

\subsection{Uniform Quantization}
\label{sec:quant}
We consider uniform per-tensor quantization of both weights and activations to $b$-bit signed values \cite{jacob}.
The set of floating-point values in an activation matrix $X$ has a range $R = max(X) - min(X)$.
Unlike weights, activation ranges vary greatly during inference depending on model inputs so an acceptable range $R$ is typically derived from activation statistics collected during training \cite{pytorch}.
To map values in $X$ to integers in $[0, 2^b - 1]$, we partition $R$ into $2^b - 1$ uniform intervals of length $s_x = \frac{R}{2^b -1}$, also called the scale factor.
For example, given an FP32 activation $x^f$, its quantized value $x^q = round(\frac{x^f}{s_x})$ maps $x^f$ into $[0, 2^b - 1]$.
Since the FP32 range of activations are asymmetric around 0 after ReLU (all positive), we shift $x^q$ by an offset $o_x = -2^{b-1} - round(\frac{min(X)}{s_x})$ into the range $[-2^{b-1}, 2^{b-1} - 1]$, guaranteeing that the FP32 value for 0 maps to an integer value, thereby:
\begin{equation} \label{eq:1}
x^q = round(\frac{x^f}{s_x}) + o_x
\end{equation}
We can obtain the approximate FP32 representation of a quantized activation $x^q$ by reversing the effect of the scale and offset via the equation:
\begin{equation} \label{eq:2}
x^{f*} = s_x(x^q - o_x)
\end{equation}
The difference $|x^f - x^{f*}|$ is the quantization error.
Each weight matrix $W$ similarly possesses a scale factor $s_w$ and offset $o_w$ derived via the same equations above either statically after training or updated during quantization-aware training (QAT).

Multiplication of an $M\times K$ weight matrix and $K\times N$ activation matrix consists of $M\cdot N$ dot products of length $K$. 
We perform dot product in the quantized domain using the approximate FP32 representations (Equation \ref{eq:2}) of weights $w$, input activations $x$, and output activations $z$
\begin{equation} \label{eq:3}
 s_z(z - o_z) = \sum_{i=1}^{K} s_w(w_i^q - o_w) s_x(x_i^q - o_x)
\end{equation}

where $s_z$ and $o_z$ represent the quantization parameters of $z$.
The FP32 scale factor terms can be factored out and normalized to an integer representation so the entire computation occurs in integer  arithmetic \cite{jacob}.
In practice, neural network weights approximate a normal distribution symmetric about zero and popular neural network libraries fix $o_w = 0$ \cite{jacob,tensorflow,termquant,pytorch}.
Then, several terms under the summation disappear and the majority of computation arises from the integer dot product $\sum_{i=1}^{K} w_i^q x_i^q$.
\begin{equation} \label{eq:3}
z = \sum_{i=1}^{K} w_i^q x_i^q
\end{equation}

\subsection{Network Pruning}
\label{sec:pruning}
Pruning is a process for reducing model size by setting low-magnitude weights to zero during training.
The resulting weight matrices are sparse and each layer's computation amounts to one or more sparse-matrix-times-dense-matrix multiplications (SpMM) between sparse weights and dense activations.
Structured pruning entails pruning away groups of weights within a network based on some criteria such as L1 or L2 norm of the group.
Since structured sparsity yields predictable nonzero distributions in weight matrices, it is amenable to software and/or hardware acceleration for skipping unnecessary computation.
However, structured pruning limits the degrees of freedom in the sparsification process and may produce models with lower accuracy than the dense baseline \cite{survey, malach}.

On the other hand, unstructured pruning entails removing individual neuron connections, producing a model with high accuracy that approaches the dense baseline. 
However, accelerating unstructured sparse matrix operations is challenging as non-zero values may be arbitrarily distributed and must be addressed individually \cite{scalpel}.
Unstructured sparse data formats, such as compressed sparse row (CSR) or compressed sparse column (CSC) formats \cite{dongarracrs} may incur significant overheads from index storage \cite{dcsr} and irregular memory accesses, especially on resource constrained systems \cite{scalpel}.
Even though the pruned network has only a fraction of the compute requirements, these overheads may cause performance of a pruned network to be worse than the corresponding dense network \cite{dcsr,scalpel}. 

In this work we focus on N:M pruning \cite{nm,nvidia_nm}, a middle-ground between structured and unstructured pruning that allows for unstructured sparsity within fixed-sized groups of weights.
In N:M pruning, the smallest N out of every M weights are pruned away and set to 0.
The resulting semi-structured sparsity is amenable to both software and hardware acceleration while incurring less compute and indexing overhead than unstructured sparse formats \cite{nvidia_nm,deepsparse}.


\section{Accumulating in Low Resolution}
\label{sec:accum}

Consider the dot product $\sum_{i=1}^{K} w_i^q x_i^q$ that arises when weights and activations are uniformly quantized to $b$ bits.
Assume we accumulate partial results into a $p$-bit register.
where each partial product $w_i^q x_i^q$ is $2b$-bits. 
This leaves $p - 2b$ bits leftover for precision during accumulation. 
Hence, our dot product may overflow when $K\geq2^{p - 2b}$.
For 8-bit quantization and a 32-bit accumulator, the threshold $K^*=2^{(32-2*8)}=65536$ is high enough to avoid overflow in most popular neural networks.
However, if we use a narrow accumulator e.g., $p = 2b$, overflows are possible after summing only 2 partial products. In order to reliably use low-resolution accumulators, we need a way of reducing such overflow.

\subsection{Characterizing Overflows}
\label{sec:ovfl}

We divide dot product overflows into two categories: \textbf{persistent} and \textbf{transient}.
A persistent overflow occurs when the final dot product result overflows regardless of the order in which partial products were added.
In other words, a persistent overflow is a true overflow where the final result is simply too large for the accumulator.
Transient overflows arise when a partial result overflows but where the final result does not actually overflow the accumulator.
They are `temporary' overflows, a direct consequence of the order of partial products accumulation.
Hence, we could potentially eliminate transient overflows by accumulating in some optimal order.

In practice, ML frameworks for quantized networks avoid overflow by either using high-precision
accumulators (e.g., 32-64 bits) or clipping partial results into a finite range (saturation arithmetic) as they are accumulated \cite{neon, cmsis1, pulpnn}.
We investigate how clipping of persistent and/or transient overflows impacts model accuracy while varying accumulator bit width.
To this end, we trained a 1-layer MLP (linear + ReLU) on the MNIST dataset \cite{mnist} with 8-bit weights and activations using QAT. 

\begin{figure}[h!]
\centering
\includegraphics[width=0.49\linewidth]{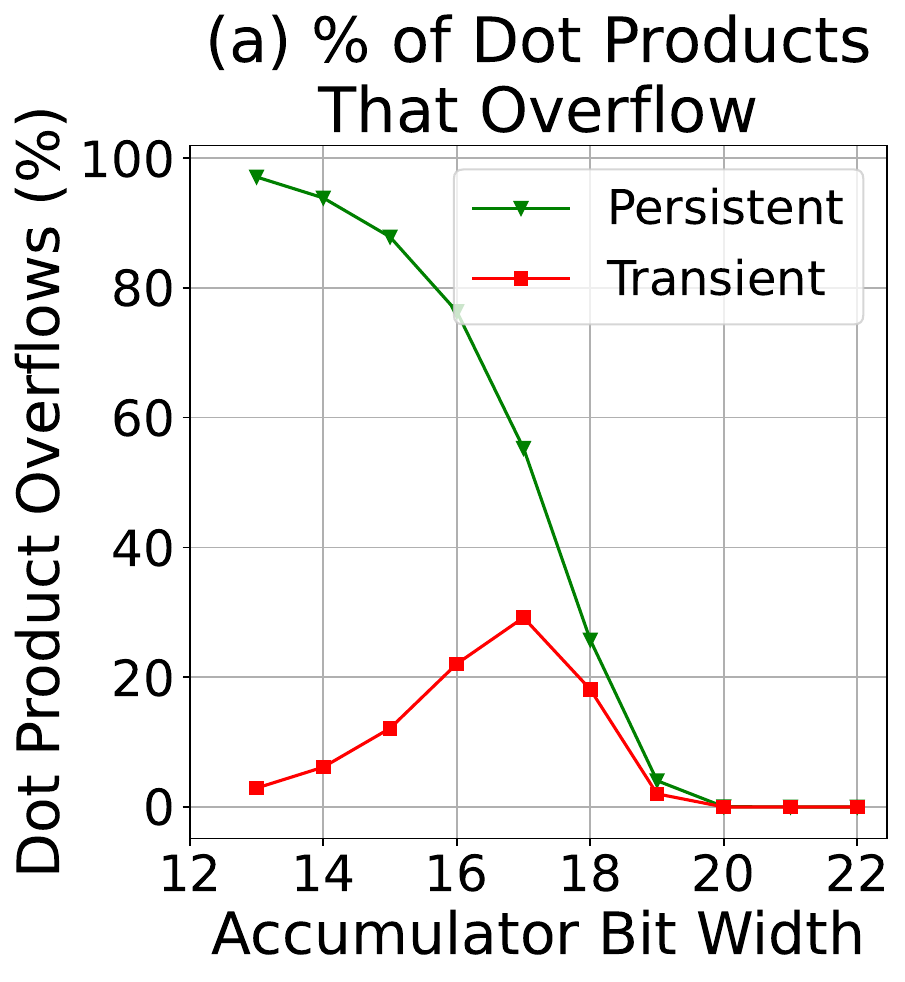}
\includegraphics[width=0.49 \linewidth]{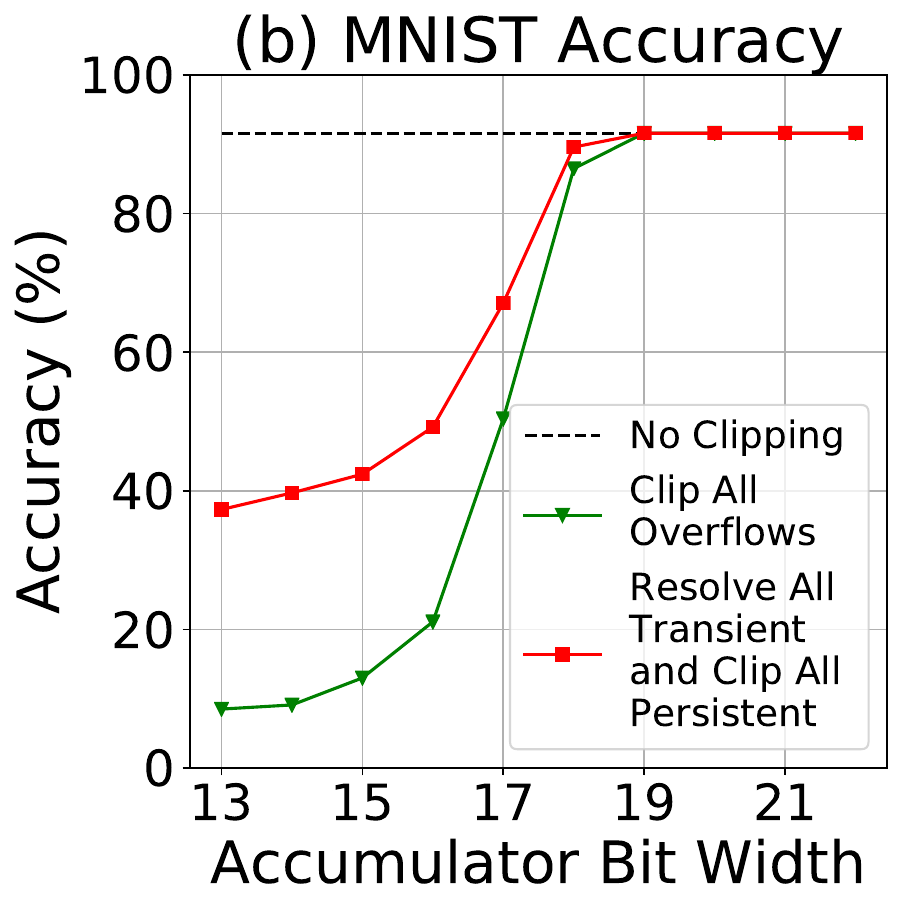}

\caption{
Profile of overflows during inference of a 1-layer MLP with 8-bit weight/activations trained on MNIST.
Even though transient overflows only account for 3\% of total overflows when using narrow 13-16 bit accumulators (a), resolving them improves accuracy from 10\% to 40\% (b) showing that for accumulators with low bitwidths, transient overflows can have a larger impact on accuracy.
}
\label{fig:mnist}
\end{figure}

Clipping overflows results in poor model accuracy when using accumulators narrower than 18 bits (Figure \ref{fig:mnist}b green).
Assume we can resolve some dot product overflows by temporarily using a high-precision accumulator for those dot products.  
Figure \ref{fig:mnist}a shows that at low resolutions of 13-16 bits, only 3-24\% of overflows are transient while the rest are persistent (97-76\%).
However, if we resolve only the transient overflows while continuing to clip all persistent overflows, accuracy improves non-trivially from ~10\% to ~40\% (Figure \ref{fig:mnist}b red).
This suggests that model accuracy becomes more sensitive to clipping transient overflows, as opposed to persistent overflows, when accumulator resolution decreases.

When we decrease the accumulator bitwidth, initially both the number of transient and persistent overflows will increase (17-20 bits). 
However, as we continue decreasing bitwidth, most overflows become persistent as the accumulator is too small to accommodate most dot products. 
As a result, the number of transient overflows decreases (17 bits or fewer). (Note that an overflow is not considered transient if the final result also overflows).
Beyond 13 bits, nearly all overflows will be persistent. 
If we prune the network, the number of terms in the dot product decreases, leading to a decrease in persistent overflows.
However, several transient overflows will still remain and degrade accuracy if not resolved (See Figure \ref{fig:accum} magenta lines).

A2Q \cite{a2q} eliminates the possibility of both transient and persistent overflows by constraining the weight vector's L1-norm during QAT. They first bound the dot product result :
\[  |\sum_{i=1}^{K} w_i^q x_i^q| \leq \sum_{i=1}^{K} |w_i^q| |x_i^q| \leq 2^{p-1} - 1\]
In the worst case, all activations are maximal $|x_i^q| = 2^{b-1}$ and the weight L1-norm may be bounded such that:
\[ \sum_{i=1}^{k} |w_i^q| = \| \mathbf{w^q}\|_1 \leq \frac{2^{p-1} - 1}{2^{b-1}}\]
This bound acts as an L1-regularizer and pulls most weight values toward zero, ensuring that partial sums never grow beyond $p$ bits.
L1 regularization promotes unstructured sparsity in the weight matrices, reducing the model size and enabling acceleration by skipping zero computations.
However, as mentioned in Section \ref{sec:pruning}, models with unstructured sparsity are more difficult to accelerate than structured sparse models on GPU, CPU, and MCU hardware.

We find that enforcing strict bounds on weight magnitude is not necessary for using narrow accumulators. Instead, we first reduce the number of weights in each dot product through N:M pruning.
Then at inference time, we algorithmically resolve transient overflows by optimally reordering the dot product, as transient overflows are simply a consequence of the order of summation.

\subsection{Sorted Dot Product}
\label{sec:sort}

In this section we present our sorted dot product algorithm for avoiding transient overflows.
Assuming the final result does not overflow the accumulator, there must exist an ordering of the summation such that no partial accumulation result overflows as well.
Transient overflows arise as a result of adding multiple large positive or large negative partial products.
We can avoid these large accumulations by first pairing the large positive and negative partial products together and summing them to cancel out their effects.
This principle is the basis for our method, detailed in pythonic pseudo-code in Algorithm \ref{algo:sort}.
In summary, our method 1) performs the pairwise product between weight and activation vectors, 2) splits the products into positive and negative arrays, 3) sort the positives descending, sort the negatives ascending, and sum the two arrays pairwise, 4) repeat step 1-3 until the final two products are summed to give the final result.

\begin{algorithm}
    \SetAlgoLined
    \SetInd{0.25em}{0.5em}
    \textbf{Input:} 
\begin{itemize}
    \item weight and activation vectors $W$ with length $K$
    \item temp storage for partial products $prods$, positive values $pos$, and negative values $neg$
\end{itemize}

    \tcp{Pairwise product of weight and activation}
    \For{$k = 0$ \KwTo $K$}{
        $prods[k] = X[k] * W[k]$
    }

    \tcp{Split, sort, and pairwise add partial products until final result is accumulated}
    \While{len(prods) > 1}{
        \tcp{Split prods into pos and neg arrays}
        $pos = prods[prods > 0]$\\
        $neg = prods[prods < 0]$\\
        \tcp{Sort pos desc, neg asc}    
        $pos = sort\_desc(pos)$\\
        $neg = sort\_asc(neg)$\\
        \tcp{Sum up the products if all same sign}    
        $m = min(len(pos), len(neg))$\\
        \lIf{$m == 0$} {\\ \quad \Return $sum(prods)$}
        \tcp{Pairwise Add pos and neg arrays}    
        $prods = []$\\
        \For{$i = 0$ \KwTo $m$}{
            $prods[i] = pos[i] + neg[i]$
        }
        \tcp{Append leftover prods that weren't paired}
        \lIf{$len(pos) > len(neg)$} {\\ \quad$prods.append(pos[m:])$}
        \lElse{\\
        \quad $prods.append(neg[m:])$}
    }
    \Return $prods[0]$\\

\caption{Sorted Dot Product}
\label{algo:sort}
\end{algorithm}

By sorting positives and negatives separately then adding them pairwise, we ensure that partial sums are accumulated in a monotonically increasing order until they reach the final result.
%
In general, a dot product between two random vectors (where the final result does not overflow) may need more than one round of positive/negative splitting followed by sorting to eliminate transient overflows.
However, neural network weights are approximately normally distributed and symmetric around 0 ensuring that pairwise products of weights and input activations (half-normally distributed after ReLU) are also symmetric.
Each dot product involves summing a roughly equal number of positive and negative pairwise products with similar range.
Hence, we can eliminate the vast majority of transient overflows through only a single sorting round.
For example, sorting dot products for a single round during MobileNetV2 inference allows for resolving 99.8\% of transient overflows.


\section{Combining Pruning and Quantization}
\label{sec:pruning}

\begin{figure}[h!]
\centering
\includegraphics[width=0.95\linewidth]{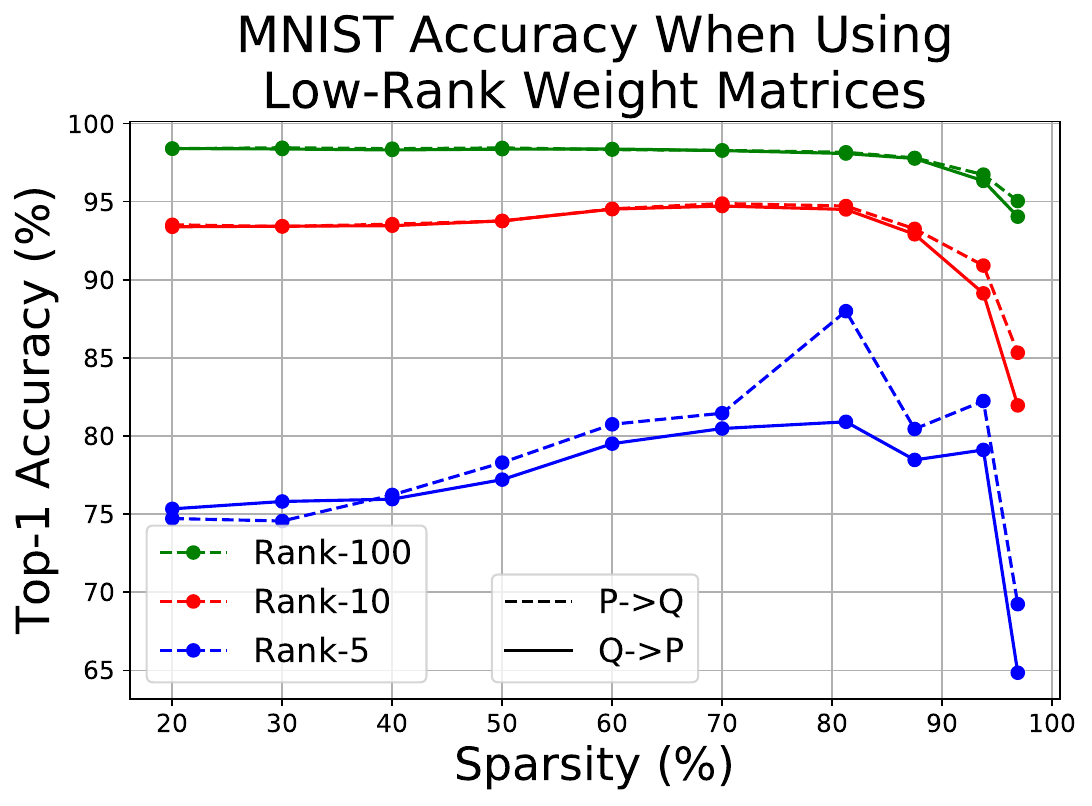}

\caption{
We compare accuracy of pruning before quantization (P->Q) with quantization before pruning (Q->P) under low-rank approximations of weight matrices in a two-layer MLP.
In contrast to Q->P, P->Q models are more resilient to low-rank weight approximations and suffer from less accuracy loss as sparsity increases.
}
\label{fig:rank}
\end{figure}

Several prior methods have successfully combined pruning and quantization to achieve high model compression rates.
For example, Deep Compression \cite{deepcc} first prunes and then quantizes weights whereas \cite{ssp} and \cite{viral} do the opposite.
Other works simultaneously schedule pruning and quantization during QAT \cite{pq,pqat} and/or jointly parameterize weight bitwidth and pruning rate in gradient updates \cite{qst,djpq}.
In essence, these methods either perform pruning in floating point before quantization or quantize to 8 (or fewer) bits before pruning low-precision weights, hereafter noted as "P->Q" and "Q->P", respectively.

While both P->Q and Q->P training methods have achieved high compression and accuracy across a variety of settings, it is unclear whether the pruned model retains more information for the subsequent classification task after FP32 pruning (via P->Q) or INT8 pruning (via Q->P).
To this end, we compare the classification accuracy of both methods when pruning a model in the presence of low-rank approximations to the weight matrix.

Before pruning a weight matrix to a target sparsity, we compute the matrix's SVD and represent the weights with a rank-$k$ approximation.
We then evaluate both methods on their ability to maintain accuracy even as weight approximations become more aggressive (i.e., as $k$ decreases).

Consider a 2-layer MLP consisting of a 784x784 hidden layer followed by a 784x10 classification head.
We train this model on MNIST for 150 epochs while performing low-rank approximation and N:M pruning (with group size $M=32$) of the hidden layer.
In P->Q, we perform FP32 training for 120 epochs while pruning weights every 10 epochs until the target sparsity (\%) is reached.
This is followed by 30 epochs of QAT to quantize the remaining weights. 
For example, to achieve 30\% sparsity, epochs 10, 20, and 30 involve a rank-$k$ approximation followed by pruning such that sparsity is $\approx$ 10\%, 20\%, and 30\% in those epochs, respectively.
In Q->P, we perform QAT for the entire 150 epochs using the pruning schedule mentioned above.
We also use the same model initialization and hyperparameters (loss function, learning rate scheduler, optimizer, etc) for both methods.

Figure \ref{fig:rank} displays accuracy of P->Q versus Q->P as we reduce the hidden layer's rank from $k=784$ to $k=\{100,10,5\}$.
At high sparsities, Q->P accuracy becomes more sensitive to low-rank approximations than P->Q.
Q->P also performs worse as rank decreases from 100 to 5.
Taken together, these observations imply that FP32 values provide a better signal for pruning than INT8.
While some joint quantization/pruning methods are able to leverage INT8 weights for pruning with minimal accuracy loss, they may require pre-training \cite{djpq} or promote hardware-unfriendly unstructured sparsity \cite{a2q}.
This further motivates us to adopt the P->Q process in generating models for low-resolution accumulation.
We evaluate our simple P->Q formulation on larger models and against prior works in Section \ref{sec:eval}.




\section{Evaluation}
\label{sec:eval}
In this section, we evaluate our PQS framework in terms of accumulator compression and model accuracy for several neural networks.
We first provide an overview of our PQS software library and network training setup.
In Section \ref{sec:pq}, we compare P->Q and Q->P training on larger models and against prior methods for combining pruning and quantization.
Then in Section \ref{sec:bit} we generate sparse models via P->Q training and compare them with prior works in terms of accuracy and accumulator bitwidth reduction.

\subsubsection{\textbf{Library for Analyzing Overflows}}
Prior works have addressed the difficulty of analyzing transient overflows due to lack of support in off-the-shelf deep learning frameworks \cite{a2q, wrapnet}.
We extend PyTorch's quantization framework with custom linear and convolution layers implementing our sorted dot product.
By fully unrolling the dot product loop in matrix computations, we are able to evaluate the end-to-end impact of sorting on reducing overflows and improving model accuracy.
We expose control of quantized dot products to the programmer allowing one to test different weight, activation, and accumulator bitwidths and evaluate other overflow solutions such as clipping or wraparound arithmetic. 
To our knowledge, our library is the first to enable fine-grained analysis of quantized dot products in neural networks.

\subsubsection{\textbf{Experiment Setup}}
We evaluate PQS using MobileNetV2 \cite{mobilenet} and ResNet-18 \cite{resnet} on CIFAR10 dataset \cite{krizhevskycifar10}.
We train these networks via the P->Q procedure to generate sparse, quantized models with N:M semi-structured sparsity.
We iteratively prune all 2D-convolution and linear layers except the first 2D convolution and final linear classifier head.
Every 10 epochs, we prune the smallest 10\% of values within each consecutive group of $M=16$ weights.
For example in epoch 10, we set the smallest 2 out of 16 ($\approx$ 10\%) values to 0 while in epoch 20, we prune such that 20\% of weights are set to 0 (3 out of every 16 values).
We uniformly prune every layer to the same sparsity (\%) and enforce the same weight, activation, and accumulator bitwidths across each layer (e.g., 8/8/16 w/act/accum bitwidths for all layers).
Once the desired sparsity is reached, we continue training the network until a total of 200 epochs have elapsed (including the pruning phase).

\subsection{P->Q vs Q->P}
\label{sec:pq}

\begin{figure}[h!]
\centering
\includegraphics[width=0.49\linewidth]{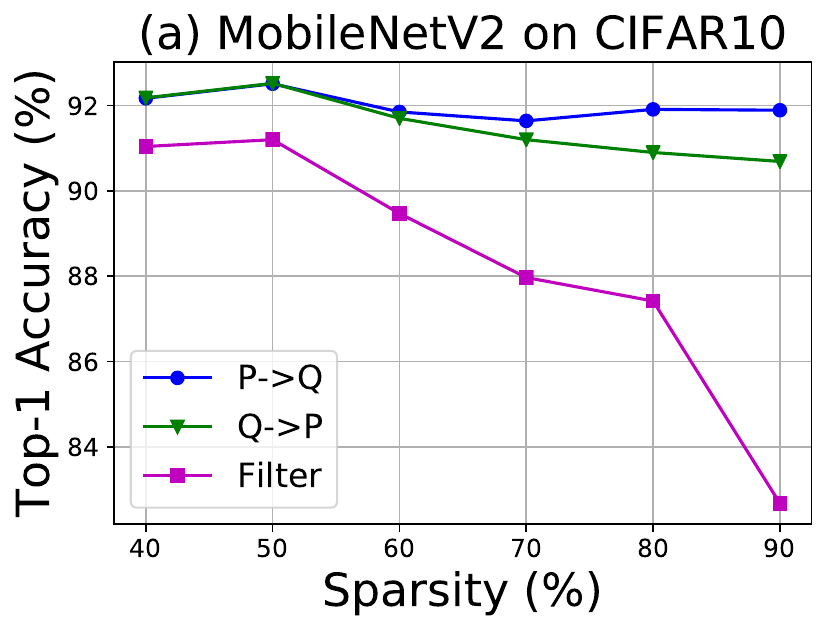}
\includegraphics[width=0.49 \linewidth]{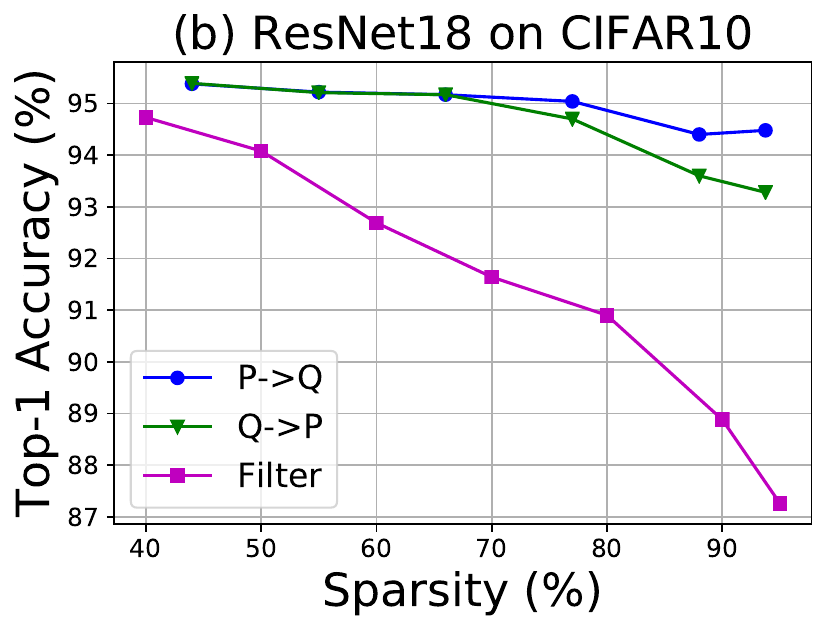}

\caption{
We compare P->Q and Q->P in MobileNetV2 (a) and ResNet-18 (b) on CIFAR10.
P->Q achieves up to 1.5\% higher accuracy than Q->P while maintaining performance at higher sparsities.
Structured filter pruning also performs poorly compared to N:M pruning in P->Q and Q->P.
}
\label{fig:pq}
\end{figure}

We evaluate the performance of P->Q and Q->P when training MobileNetV2 and ResNet-18.
We follow similar training procedures as in Section \ref{sec:pruning} except we keep weight matrices in full rank.
For P->Q, we train in floating point for 180 epochs followed by 20 epochs of QAT whereas for Q->P, we pruning every 10 epochs during 200 epochs of QAT.

In Figure \ref{fig:pq} we observe similar trends for MobileNetV2 and ResNet-18 as in the previous case of our two-layer MLP on MNIST.
When running P->Q using filter pruning instead of N:M pruning (magenta line), accuracy degrades significantly.

\subsection{Reducing Accumulator Bitwidth}
\label{sec:bit}

In this section, we evaluate the ability of PQS to enable use of narrow accumulators while maintaining FP32 model accuracy in ResNet-18 and MobileNetV2.
We sweep the PQS design space by training several P->Q models with varying sparsity and data bitwidths i.e., weight, activation, and accumulator bitwidth.
We vary weight and activations from 5 to 8 bits while varying the accumulator from 12 to 24 bits.
We select the best performing models with the lowest required accumulator bitwidth to generate a pareto frontier.
For models on the frontier, we use our software library to evaluate the accuracy impact when we clip instead of sort the dot product accumulations.

\begin{figure}[h!]
\centering
\includegraphics[width=0.49\linewidth]{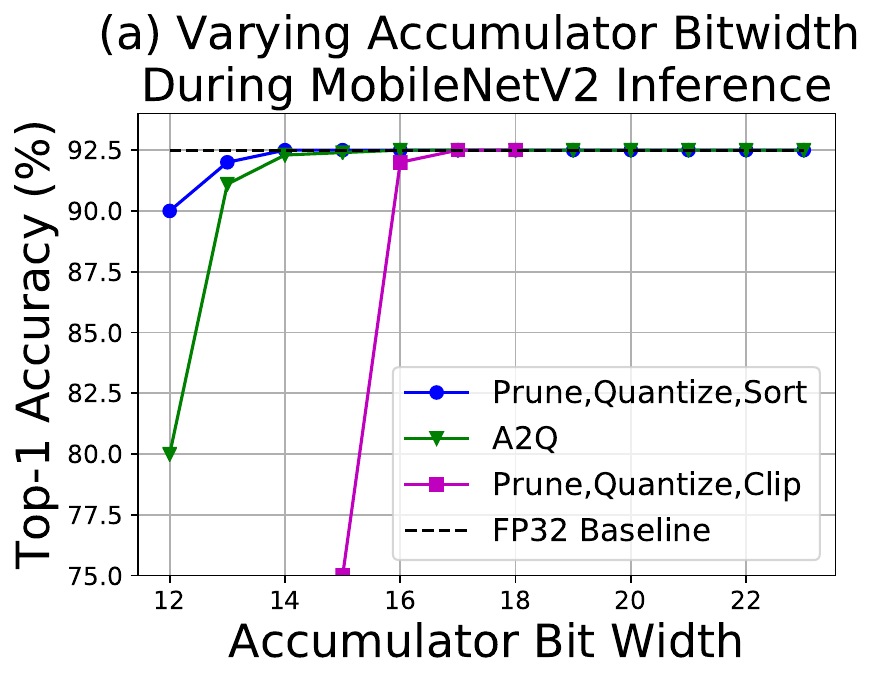}
\includegraphics[width=0.48 \linewidth]{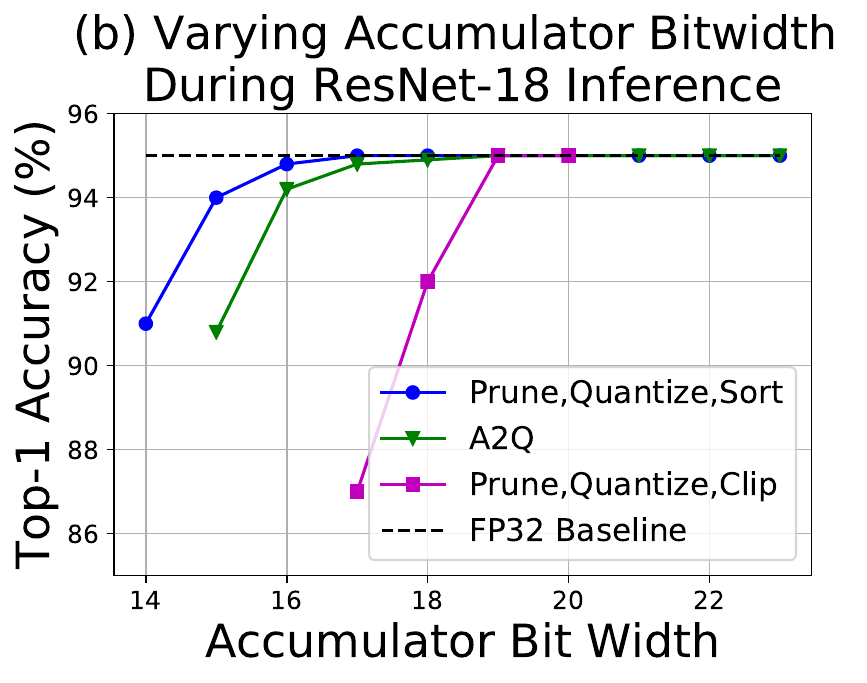}

\caption{
We visualize the trade-off between accumulator bit width and model accuracy. 
PQS (blue) can make use of accumulators with lower bitwidth than A2Q, without sacrificing significant model performance relative to the floating-point baseline.
Sorting before accumulating the dot product allows us to avoid transient overflows and use up to 4 fewer bits in the accumulator than if we clipped those overflows (magenta lines).
}
\label{fig:accum}
\end{figure}

Figure \ref{fig:accum} shows that PQS can push the accumulator bit width lower than A2Q while also maintaining task performance.
Models on the PQS frontier are roughly 80-95\% sparse.
While pruning can reduce the length of individual dot products, transient overflows still arise during inference.
The magenta lines show that clipping transient overflows within sparse dot products can limit how much we may reduce accumulator bitwidth.
Sorting before accumulating allows us to avoid transient overflows and reduce accumulator resolution by $\approx$ 4 bits while maintaining model accuracy.

\section{Discussion}
So far, we have presented a framework for reducing accumulator bitwidth that improves over the current state-of-the-art methods.
In this section, we consider the benefits and limitations of our method in the context of hardware/software implementations. 

\textbf{Structured Sparsity} Matrices pruned with N:M sparsity have a lower memory footprint than those pruned with unstructured sparsity.
Quantization itself induces additional sparsity after pruning. 
This is because weights are normally distributed around zero, hence most weights are small and get placed in to the 0 bucket after uniform quantization.
Since the weights were already pruned with N:M sparsity, the additional sparsity from quantization creates whole blocks of zeros.
We can skip computation of whole blocks of zeros for further acceleration in PQS.

\textbf{Software Scheduling}: Weight and activation matrices are typically tiled (partitioned) into blocks and multiplied block-wise in the cache to increase data reuse and reduce external memory bandwidth usage \cite{blis,cake}.
Tiling splits a single dot product into multiple independent dot products.
However, our dot product algorithm requires all partial products to be computed before sorting to eliminate transient overflows.
Hence, our algorithm as presented is not compatible with tiling and would introduce significant overhead when sorting longer dot products e.g., 4000 partial sums in certain transformer models.
We compared the ability of sorting the full dot product versus sorting tiled dot products to mitigate transient overflows.
For example, with a tile size of $k=256$, we are still able to eliminate 99\% of all transient overflows in MobileNetV2 inference via PQS.

\textbf{Hardware Implementation}:
Sorting networks such as the bitonic algorithm are popular for sorting arrays in hardware \cite{ssn,fsort}.
Such networks compare and swap elements in an array using multiple comparators in parallel across several pipeline stages.
Since tiling the dot product shortens the array we need to sort, the sorting network needs fewer comparators and stages, saving area and power.

By sorting partial products before adding, we transform the original accumulation sequence into a monotonic one where the accumulating partial sum always increases.
Hence, if the accumulator does overflow, the rest of the partial sum is guaranteed to overflow.
This allows us to identify persistent overflows earlier in the dot product so we may clip the partial result and avoid any further accumulations.
We plan on investigating the hardware implications of PQS in future work.

\section{Conclusion}
Pruning and quantization are two major techniques that are typically used sequentially in reducing model size, but in the prior literature, they are normally analyzed separately. 
In this work, we analyze how pruning followed by quantization, or vice versa, can avoid accumulation overflows in computing dot products.

We show that FP32 weights, before being quantized, present high-quality signals in determining which model weights to be pruned. That is, pruning followed by quantization can reduce persistent overflows in dot product computation. 
Furthermore, during inference, if we sum partial products in a sorted manner, we can minimize transient overflows. These techniques together form the proposed PQS (Prune, Quantize, and Sort) method of this paper in achieving low-bitwidth accumulation of dot products in neural network computations. To the best of our knowledge, the PQS method and our analysis are novel in the literature.

\bibliographystyle{ACM-Reference-Format}
\bibliography{refs}


\end{document}